\pdfoutput=1

\documentclass[11pt]{article}

\usepackage[preprint]{acl}

\usepackage{times}
\usepackage{latexsym}

\usepackage[T1]{fontenc}

\usepackage[utf8]{inputenc}

\usepackage{microtype}

\usepackage{inconsolata}

\usepackage{graphicx}

\usepackage{algorithm}
\usepackage{algorithmicx}
\usepackage[noend]{algpseudocode}
\usepackage{amsmath}
\usepackage{amssymb}
\usepackage{array}
\usepackage{arydshln}
\usepackage{bm}
\usepackage{booktabs}
\usepackage{color}
\usepackage{colortbl}
\usepackage{enumitem}
\usepackage{fancyvrb}
\usepackage[T1]{fontenc}
\usepackage{graphicx}
\usepackage{multirow}
\usepackage{pifont}
\usepackage{soul}
\usepackage{stfloats}
\usepackage{subcaption}
\usepackage{xcolor}

\newcommand{\tightparagraph}[1]{\vspace{-4pt} \paragraph{#1}}

\definecolor{difcolor}{RGB}{0,0,192}

\newcommand{\methodname}{\textsc{OLMoTrace}}
\newcommand{\infinigram}{infini-gram}
\newcommand{\Infinigram}{Infini-gram}

\title{\methodname{}: Tracing Language Model Outputs \\Back to Trillions of Training Tokens}

\newcommand{\aspace}{\hspace{0.7em}}
\newcommand{\aitwo}{$^{\alpha}$}
\newcommand{\uw}{$^{\omega}$}
\newcommand{\berkeley}{$^{\beta}$}

\newcommand{\stanford}{$^{\sigma}$}
\author{
  Jiacheng Liu\aitwo \uw \aspace 
  Taylor Blanton\aitwo \aspace 
  Yanai Elazar\aitwo \uw \aspace 
  Sewon Min\aitwo \berkeley \vspace{3pt} \\ 
  \textbf{YenSung Chen}\aitwo \aspace 
  \textbf{Arnavi Chheda-Kothary}\aitwo \uw \aspace 
  \textbf{Huy Tran}\aitwo \aspace 
  \textbf{Byron Bischoff}\aitwo \aspace 
  \textbf{Eric Marsh}\aitwo \vspace{-1pt} \\ 
  \textbf{Michael Schmitz}\aitwo \aspace 
  \textbf{Cassidy Trier}\aitwo \aspace 
  \textbf{Aaron Sarnat}\aitwo \aspace 
  \textbf{Jenna James}\aitwo \aspace 
  \textbf{Jon Borchardt}\aitwo \vspace{-1pt} \\ 
  \textbf{Bailey Kuehl}\aitwo \aspace 
  \textbf{Evie Cheng}\aitwo \aspace 
  \textbf{Karen Farley}\aitwo \aspace 
  \textbf{Sruthi Sreeram}\aitwo \aspace
  \textbf{Taira Anderson}\aitwo \vspace{-1pt} \\ 
  \textbf{David Albright}\aitwo \aspace 
  \textbf{Carissa Schoenick}\aitwo \aspace 
  \textbf{Luca Soldaini}\aitwo \aspace 
  \textbf{Dirk Groeneveld}\aitwo \\ 
  \textbf{Rock Yuren Pang}\uw \vspace{3pt} \\ 
  \textbf{Pang Wei Koh}\aitwo \uw \aspace 
  \textbf{Noah A. Smith}\aitwo \uw \aspace 
  \textbf{Sophie Lebrecht}\aitwo \aspace 
  \textbf{Yejin Choi}\stanford \vspace{-1pt} \\ 
  \textbf{Hannaneh Hajishirzi}\aitwo \uw \aspace 
  \textbf{Ali Farhadi}\aitwo \uw \aspace 
  \textbf{Jesse Dodge}\aitwo \vspace{3pt} \\ 
  \aitwo{}Allen Institute for AI \aspace
  \uw{}University of Washington \aspace
  \berkeley{}UC Berkeley \aspace
  \stanford{}Stanford University
}

\begin{document}

\maketitle


\begin{abstract}
\vspace{-2pt}
We present \methodname{}, the first system that traces the outputs of language models back to their full, multi-trillion-token training data in real time.
\methodname{} finds and shows verbatim matches between segments of language model output and documents in the training text corpora.
Powered by an extended version of \infinigram{} \citep{Liu2024InfinigramSU}, our system returns tracing results within a few seconds.
\methodname{} can help users understand the behavior of language models through the lens of their training data.
We showcase how it can be used to explore fact checking, hallucination, and the creativity of language models.
\methodname{} is publicly available and fully open-source. 
\end{abstract}

\begin{figure*}[!b]
\centering
\vspace{-12pt}
\includegraphics[width=0.68 \textwidth]{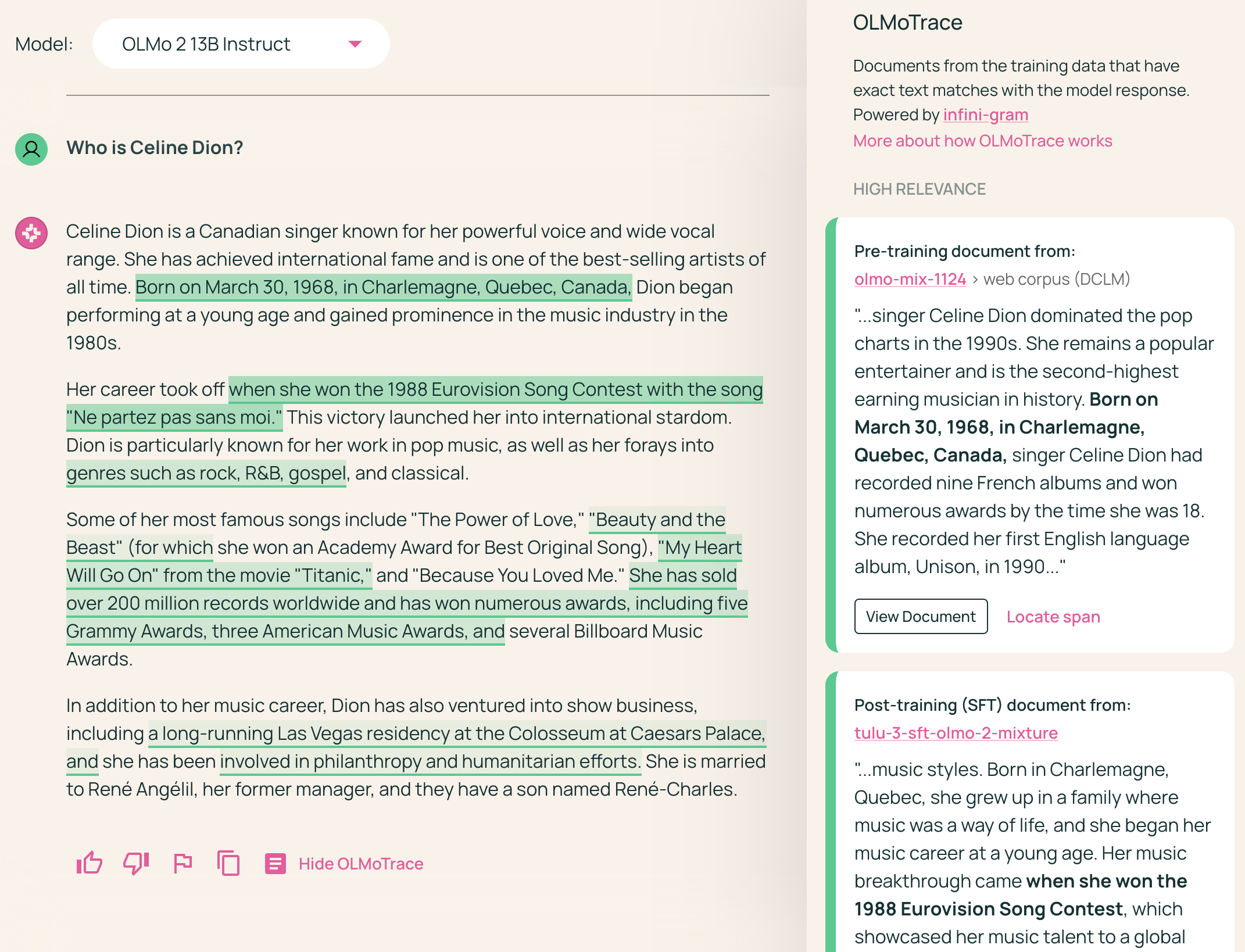}
\hfill
\includegraphics[width=0.305 \textwidth]{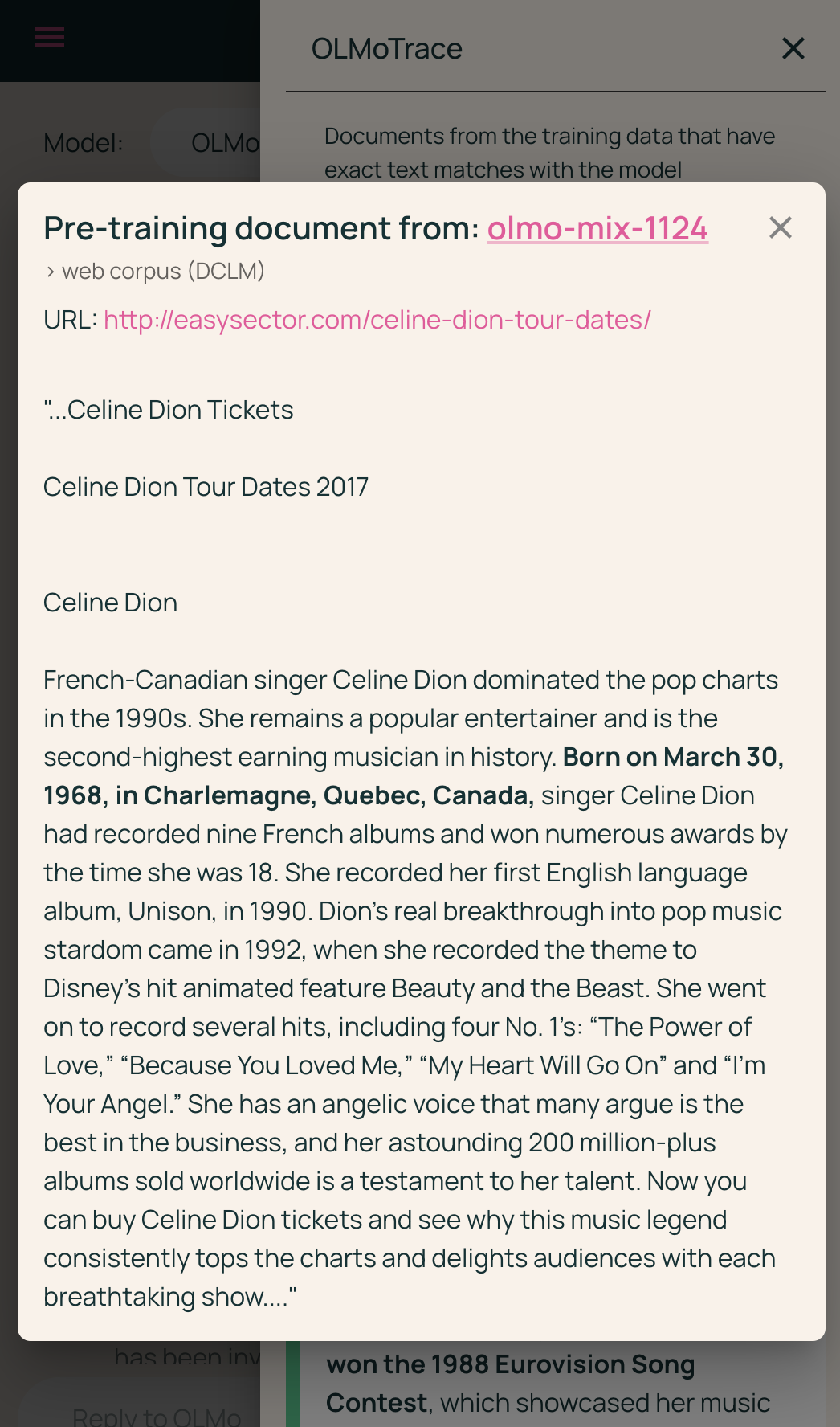}
\vspace{-7pt}
\caption{
    \methodname{} on Ai2 Playground.
    \textbf{Left:} On a response generated by OLMo, \methodname{} highlights text spans found verbatim in the model's training data and shows their source documents.
    Brighter highlights indicate spans from more relevant training documents, while darker highlights denote less relevant ones.
    \textbf{Right:} When user clicks the ``View Document'' button, the document is shown with extended context.
    Try \methodname{} at \url{https://playground.allenai.org}.
}
\label{fig:playground}
\end{figure*}

\section{Introduction}
\label{sec:intro}

Tracing the outputs of language models (LMs) back to their training data is an important problem.
As LMs gain adoption in higher-stakes scenarios, it is critical to understand \textit{why} they generate certain responses.
However, these modern LMs are trained on massive text corpora with trillions of tokens, which are often proprietary. Fully open LMs (e.g., OLMo; \citealt{OLMo20242O2}) enable access to the training data, but  existing behavior tracing methods \citep{Koh2017UnderstandingBP,Khalifa2024SourceAwareTE,Huang2024TrainingLM} have not been scaled to work within this multi-trillion-token setting due to their heavy computational needs.

In this paper, we introduce \methodname{}, a system that traces LM outputs \textit{verbatim} back to its \textbf{full} training data and displays the tracing results to LM users in \textbf{real time}. 
Given an LM response to a user prompt, \methodname{} retrieves documents from the model's training data that contain exact matches with pieces of the LM response that are long, unique, and relevant 
to the whole response; see \autoref{fig:playground} for an example. 

The key idea that makes \methodname{} fast is that exact matches can be quickly located in a large text corpus if we pre-sort all of its suffixes lexicographically. 
We use \infinigram{} \citep{Liu2024InfinigramSU} to index the training data and develop 
a novel parallel algorithm to speed up the computation of matching spans (\S\ref{sec:pipeline}).
In our production system, \methodname{} completes tracing for each LM response (avg. $\sim$450 tokens) within 4.5 seconds on average. 

The purpose of \methodname{} is to give users a tool to explore where LMs \textit{may} have learned to generate certain word sequences,
focusing on verbatim matching as the most direct connection between LM outputs and the training data. 
\methodname{} offers an interactive experience, so that users can explore which training documents contain a specific span in the LM response, or inspect a particular document and locate its matching spans in the LM response. 
We present three case studies for ways to use \methodname{} (\S\ref{sec:usage}): (1) fact checking, 
(2) tracing the LM-generated ``creative'' expressions, 
and (3) tracing math capabilities.
We invite the community to explore more use cases to better understand the relationship between data and models.

\methodname{} is available in the Ai2 Playground\footnote{\url{https://playground.allenai.org}} and supports the three flagship OLMo models \citep{OLMo20242O2,Muennighoff2024OLMoEOM} including OLMo-2-32B-Instruct.\footnote{\scriptsize \url{https://huggingface.co/allenai/OLMo-2-0325-32B-Instruct}}
For each model, it matches against its full training data, including pre-training, mid-training, and post-training. 
\methodname{} can be applied to any LM as long as the service provider has access to its full training data.
The core part of the system is open-sourced under the Apache 2.0 license.\footnote{\url{https://github.com/allenai/infinigram-api}} 
\section{System Description}
\label{sec:system}




\tightparagraph{Features of \methodname{}.}
\autoref{fig:playground} shows \methodname{} applied to an LM response.
When \methodname{} is enabled in the Ai2 Playground, 
it highlights the matching spans in the response, and shows all training documents matching at least one of these spans in a document panel.
\methodname{} supports inspecting the documents that match with any particular highlighted span (App.~\autoref{fig:playground_more}, left), and locating the spans enclosed by any particular document (App.~\autoref{fig:playground_more}, right).
In the document panel, each document is shown with a snippet of 80 tokens surrounded the matched span; \methodname{} allows users to further inspect the document with an extended context (500 tokens).


\begin{table}
\centering
\setlength{\tabcolsep}{4pt}
\resizebox{\linewidth}{!}{%
\begin{tabular}{l l r r}
\toprule
\textbf{Stage} & \textbf{Dataset} & \textbf{\# Docs} & \textbf{\# Tokens} \\
\midrule
pre-training & \href{https://huggingface.co/datasets/allenai/olmo-mix-1124}{allenai/olmo-mix-1124} & 3081 M & 4575 B \\
mid-training & \href{https://huggingface.co/datasets/allenai/dolmino-mix-1124}{allenai/dolmino-mix-1124} & 81 M & 34 B \\
post-training & \href{https://huggingface.co/datasets/allenai/tulu-3-sft-olmo-2-mixture-0225}{SFT} \& \href{https://huggingface.co/datasets/allenai/olmo-2-0325-32b-preference-mix}{DPO} \& \href{https://huggingface.co/datasets/allenai/RLVR-GSM-MATH-IF-Mixed-Constraints}{RLVR} & 1.7 M & 1.6 B \\
\midrule
\textbf{Total} & & \textbf{3164 M} & \textbf{4611 B} \\
\bottomrule
\end{tabular}
}%
\vspace{-7pt}
\caption{
    The full training data of OLMo-2-32B-Instruct, which \methodname{} matches against. 
    For mid-training data, we excluded sources that already appeared in the pre-training data, from both the statistics and the index. 
}
\vspace{-12pt}
\label{tab:corpus_stats}
\end{table}

\tightparagraph{The training data.}
The three supported OLMo models are trained on the same pre-training and mid-training data, and slightly different post-training data.
\methodname{} matches against the entirety of an LM's training data.
\autoref{tab:corpus_stats} shows links and statistics of the training data of OLMo-2-32B-Instruct, which totals 3.2 billion documents and 4.6 trillion (Llama-2) tokens.
The other two OLMo models have similar training data size.

\section{The Inference Pipeline}
\label{sec:pipeline}

\begin{figure*}[!t]
\includegraphics[width=\textwidth]{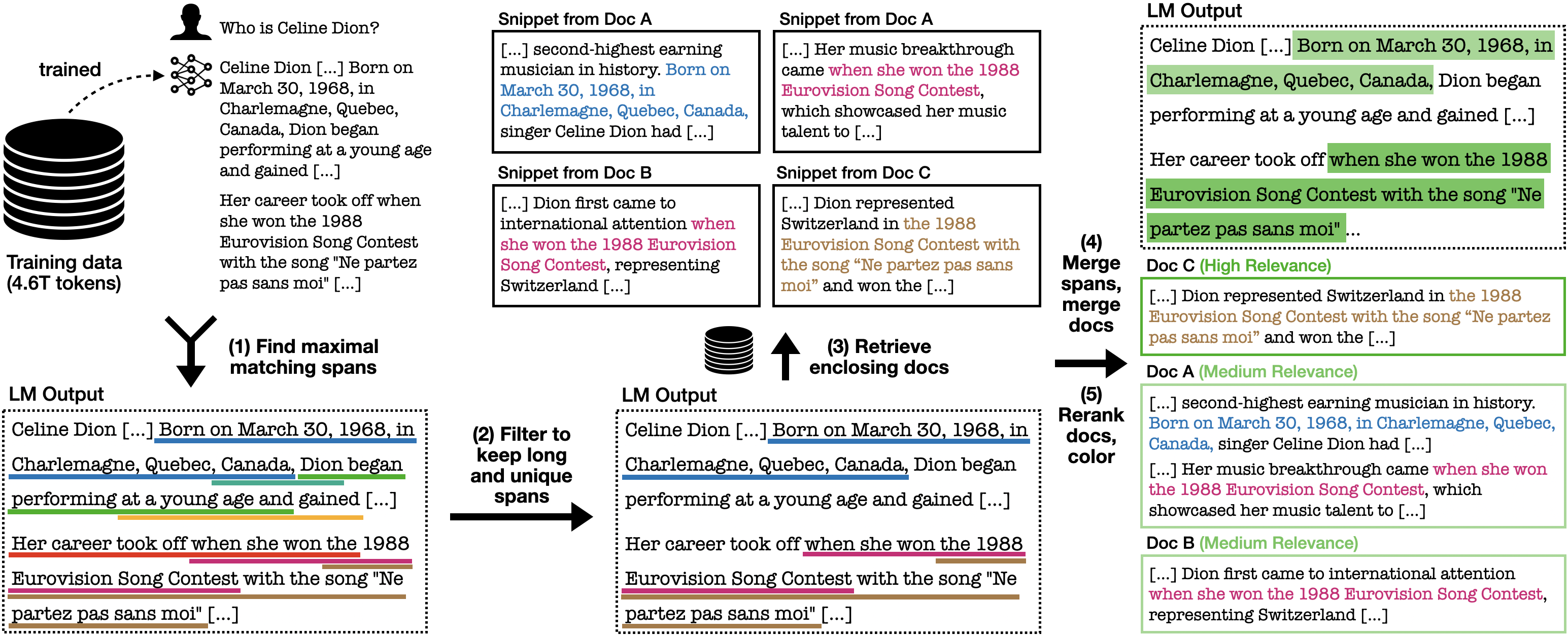}
\vspace{-8pt}
\caption{
    The \methodname{} inference pipeline, as described in \S\ref{sec:pipeline}.
    For better illustration, we slightly adjusted the highlighted spans and document relevance from the actual example.
}
\vspace{-12pt}
\label{fig:pipeline}
\end{figure*}

\methodname{} takes as input an LM response to a user prompt, and outputs (1) a set of text spans in the LM response, each marked by its starting and ending position, and (2) a list of documents from the training data of this LM, each containing one or more of the aforementioned text spans.
The \methodname{} inference pipeline consists of the following five steps (illustrated in \autoref{fig:pipeline}):

\tightparagraph{Step 1: Find maximal matching spans.}
We find all maximal spans in the LM output that appear verbatim in the training data.
Specifically, we first tokenize the LM output with the Llama-2 tokenizer, and find all spans of the token ID list that satisfy the following criteria:
\begin{enumerate}[leftmargin=16pt, itemsep=-4pt, topsep=4pt]
\item \textbf{Existence:} The span appears verbatim at least once in the training data;
\item \textbf{Self-contained:} The span does not contain a period token (\texttt{.}) or newline token (\texttt{\textbackslash{}n}) unless it appears at the end of the span; and the span does not begin or end with incomplete words;
\item \textbf{Maximality:} The span is not a subspan of another span that meets the above two criteria.
\end{enumerate}
This is the most compute-heavy step, since naively we need to enumerate all $O(L^2)$ spans of the LM output (where $L$ is the length of the LM output in tokens, and typically $L \in [10^2, 10^3]$) and scan the entire training data (with $N$ tokens where $N > 10^{12}$).
We propose a fast algorithm to compute these maximal matching spans (\S\ref{sec:algo}), which reduces the time complexity to $O(L \log N)$, and latency to $O(\log N)$ when fully parallelized.
After this step, we have a set of relatively long spans that appear in the training data.

\tightparagraph{Step 2: Filter to keep long and unique spans.}
To declutter the UI and only show spans that are more likely ``interesting'', we filter spans to keep ones with the smallest \textit{span unigram probability}, a metric that captures both length and uniqueness.
The span unigram probability is defined as the product of unigram probabilities of all tokens in the span, where the token unigram probability derived from statistics of the LM's entire training data. 
(We pre-compute and cache the token unigram probability for the entire vocabulary.)
A lower span unigram probability usually means the span is relatively long and contains non-common tokens. 
We keep $K$ spans with the smallest unigram probability, where $K = \lceil 0.05 \times L \rceil$.

During development, we tried keeping the longest spans instead of those with smallest span unigram probability.
However, we found that ranking with the span length metric leads to worse relevance level on documents retrieved from the filtered spans (see measurement of relevance in App.~\S\ref{sec:relevance_more} and \autoref{tab:relevance_eval}), 
and thus we favored the span unigram probability metric.
We chose unigram over bigram or trigram because they computing them (either online or pre-caching) takes a lot of time. 

\tightparagraph{Step 3: Retrieve enclosing documents.}
For each kept span, we retrieve up to 10 document snippets from the training data that enclose this span.
Due to the maximality criterion in step 1, most spans appear no more than 10 times. If a span exceeds this limit, we randomly sample 10 to keep retrieval time manageable and avoid UI overload.

\tightparagraph{Step 4: Merge spans, merge documents.}
To further declutter the UI, we merge (i.e., take the union of) overlapping spans into a single span to be highlighted in the LM output. 
Also, if two snippets are retrieved from the same document, we merge them into a single document to be displayed in the document panel.

\tightparagraph{Step 5: Rerank and color documents by relevance.}
To prioritize showing the most relevant documents, in the document panel we rank all documents by a BM25 score in descending order.
The per-document BM25 score is computed by treating the collection of retrieved documents as a ``corpus'', and the concatenation of user prompt and LM response as the ``query''.\footnote{We use the implementation in \url{https://github.com/dorianbrown/rank_bm25}}
We use this BM25 score because it has fairly high agreement with human judgment on topical relevance (\S\ref{sec:relevance}), and can be quickly computed using CPUs.
Subsequently, we bucket the BM25 scores into three levels -- high relevance, medium relevance, and low relevance -- and display a colored sidebar on each document to represent its relevance level.
High relevance are highlighted with the most saturated color, and low relevance with the least saturated color.
We also apply this differential coloring on span highlights: a span's relevance level is computed as the maximum relevance level among documents enclosing the span.
As a result, users are more likely to find highly relevant documents for spans highlighted with the most saturated color.

\subsection{Fast Span Computation}
\label{sec:algo}

\begin{figure*}[!t]
\includegraphics[width=\textwidth]{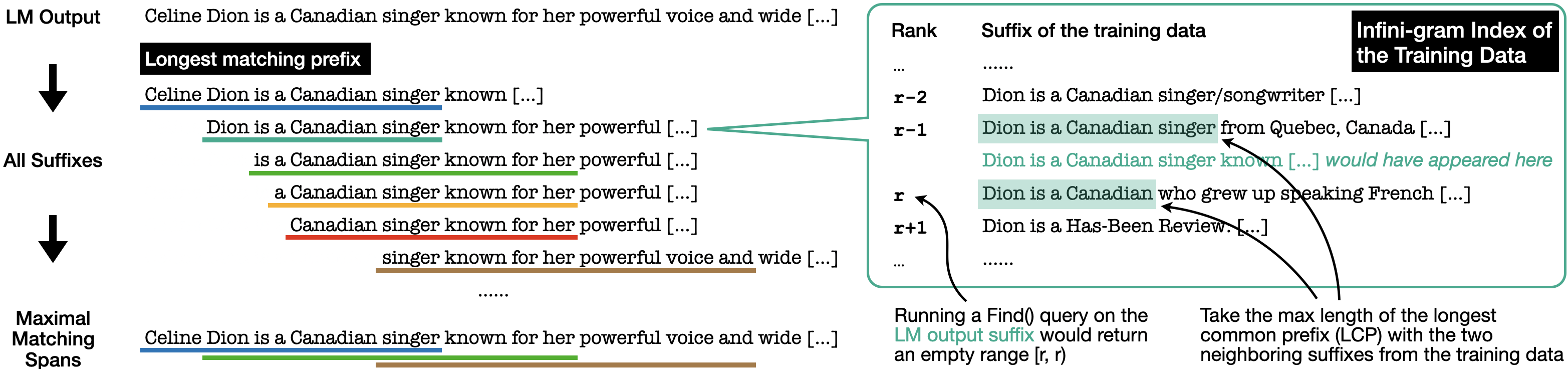}
\vspace{-8pt}
\caption{
    Computation of the maximal matching spans (\S\ref{sec:algo}).
    For each suffix of the LM output, \methodname{} computes its longest matching prefix (color-underlined) with a single \textsc{Find} query on the \infinigram{} index of the LM training data.
    All suffixes of the LM output are processed in parallel.
    Finally, non-maximal spans are suppressed.
}
\vspace{-12pt}
\label{fig:span}
\end{figure*}

Efficiently identifying all maximal matching spans across multi-trillion-token corpora is a non-trivial challenge.
To tackle this, we index the training corpora with \infinigram{} \citep{Liu2024InfinigramSU} and develop a new parallel algorithm for fast span computation.

\tightparagraph{\Infinigram{}.}
\Infinigram{} is a text search engine.
It supports efficiently counting text queries and retrieving matching documents in massive text corpora with trillions of tokens.
To make operations fast, \infinigram{} indexes text corpora with the suffix array (SA) data structure, and at inference time keeps the huge index files on low-latency SSD disks to avoid loading them into RAM.
For \methodname{}, we build an \infinigram{} index on the tokenized version of the LMs' training data (using the Llama-2 tokenizer).
On top of this index, in this work we devise a novel parallel algorithm to compute maximal matching spans with low compute latency (\autoref{fig:span} and Algorithm \ref{alg:algo}); 
we discuss this algorithm and its implementation below.

\begin{algorithm}[!t]
\footnotesize
\caption{Compute maximal matching spans.}
\textbf{Input} Model output $S_{1:L}$ (tokenized), training text corpus $T_{1:N}$ (tokenized) and its suffix array $A_{1:N}$
\begin{algorithmic}
\Procedure{GetMaximalMatchingSpans}{$S, T, A$}
    \State spans $\gets []$
    \For{$b$ = 1, \ldots, $L$} \Comment{execute in parallel}
        \If{$S_b$ is a begin-of-word token}
            \State $len \gets$ \Call{GetLongestPrefixLen}{$S_{b:L}, T, A$}
            \State spans $\gets$ spans + $[(b, b + len)]$
        \EndIf
    \EndFor
    \State \Return \Call{SuppressNonMaximalSpans}{spans}
\EndProcedure
\Procedure{GetLongestPrefixLen}{$s, T, A$}
    \State $(l, r) \gets$ \Call{Find}{$s, T, A$} \Comment{an \infinigram{} query}
    \If{$l \ne r$} \Comment{non-empty segment, $s$ is found in $T$}
        \State $len \gets |s|$
    \Else \Comment{empty segment, $s$ is not found in $T$}
        \State $len1 \gets$ \Call{LongestPrefixLen}{$s, T_{A[l]:}$}
        \State $len2 \gets$ \Call{LongestPrefixLen}{$s, T_{A[l+1]:}$}
        \State $len \gets \max(len1, len2)$
    \EndIf
    \While{$s_{:len-1}$ contains a delimiter token OR $s_{len+1}$ is not a begin-of-word token}
        \State $len \gets len - 1$
    \EndWhile
    \State \Return $len$
\EndProcedure
\Procedure{SuppressNonMaximalSpans}{spans}
    \State sort spans by beginning position in ascending order
    \State newspans $\gets []$
    \State maxend $\gets$ 0
    \For{$(b, e)$ in spans}
        \If{maxend $< e$}
            \State maxend $\gets e$
            \State newspans $\gets$ newspans + $[(b, e)]$
        \EndIf
    \EndFor
    \State \Return newspans
\EndProcedure
\end{algorithmic}
\label{alg:algo}
\end{algorithm}

\tightparagraph{Problem analysis.}
The problem of finding all maximal matching spans can be broken down into two steps: (1) finding the longest matching prefix of each suffix of the LM output; 
and (2) suppressing the non-maximal spans.
This is because starting from each position, there can be at most one span that is a maximal matching span (if there are two, then one is a subspan of the other and thus is not maximal).
The first step consists of multiple independent tasks that can be parallelized, and as we will show below, each task can be done with one \textsc{Find} query. 
\textsc{Find} is a core query operation in \infinigram{}; it returns the segment of SA that corresponds to all occurring positions of a search term in the text corpus.
Since in \infinigram{}, the processing speed of \textsc{Find} queries is bounded by disk I/O latency and there is a lot of unused throughput, parallelizing these queries can reduce the overall compute latency.
In fact, with parallelization, the overall processing speed is bottlenecked by the disk I/O throughput, and thus in our production system we store the index files on high-IOPS SSD disks. 

\tightparagraph{Finding the longest matching prefix of a suffix.}
With \textsc{Find} queries, the length of the outputted segment is the count of the search term in the text corpus.
Naively, we can run \textsc{Find} queries on incrementally long prefixes of the LM output's suffix until the count becomes zero (which takes $O(L)$ queries), or we can do binary-lifting + binary-search to reduce to $O(\log L)$ queries.
However, we show below that we can do this with one single \textsc{Find} query ($O(1)$).

We use the fact that when the search term does not exist in the text corpus, \textsc{Find} 
would return a 0-length segment (delimited by a left-inclusive starting position and a right-exclusive ending position that are identical), where the previous (or next) SA element corresponds to the suffix in the text corpus that lexicographically precedes (or succeeds) the search term (see \autoref{fig:span}).
Consequently, the suffix in the text corpus that shares the longest common prefix (LCP) with the search term must come from one of these two neighboring suffixes, and inspecting these two suffixes would tell us the length of the longest matching prefix for this search term.
Therefore, we can simply run \textsc{Find} once with the entire LM output's suffix to find out its longest matching prefix. 

In reality, we shard the \infinigram{} index because each shard is limited to 500B tokens.
In case there are multiple shards, we run \textsc{Find} on each one in parallel, and take the maximum of LCP length from all shards. 

Note that to retrieve documents containing the longest matching prefix, we need to run a second \textsc{Find} query to locate all its occurrences in the SA.
In practice, we run this query immediately after the first one to leverage temporal locality in the disk cache.

\tightparagraph{Suppressing non-maximal spans.}
We gather the longest matching prefix of all suffixes into a list of spans.
These spans begin at monotonically increasing positions, but end at monotonically non-decreasing positions that may still be identical, and thus there may still be non-maximal spans (see \autoref{fig:span}). 
To remove the non-maximal spans, we make a pass on the spans in increasing order of the beginning position, and only keep spans with an ending position larger than that of the previously encountered spans.

\begin{figure*}[!t]
\centering
\includegraphics[width=0.30 \textwidth]{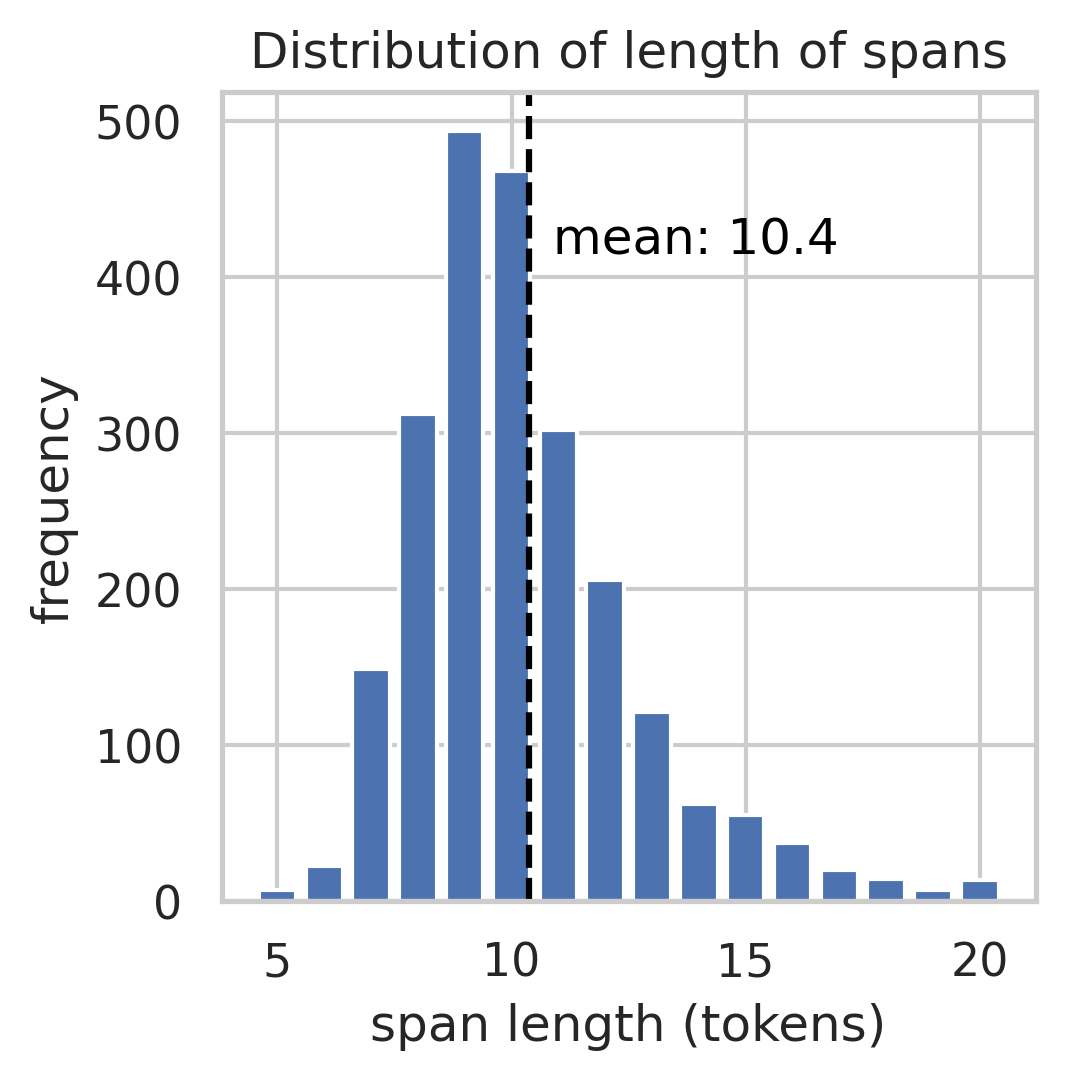}
\hfill
\includegraphics[width=0.34 \textwidth]{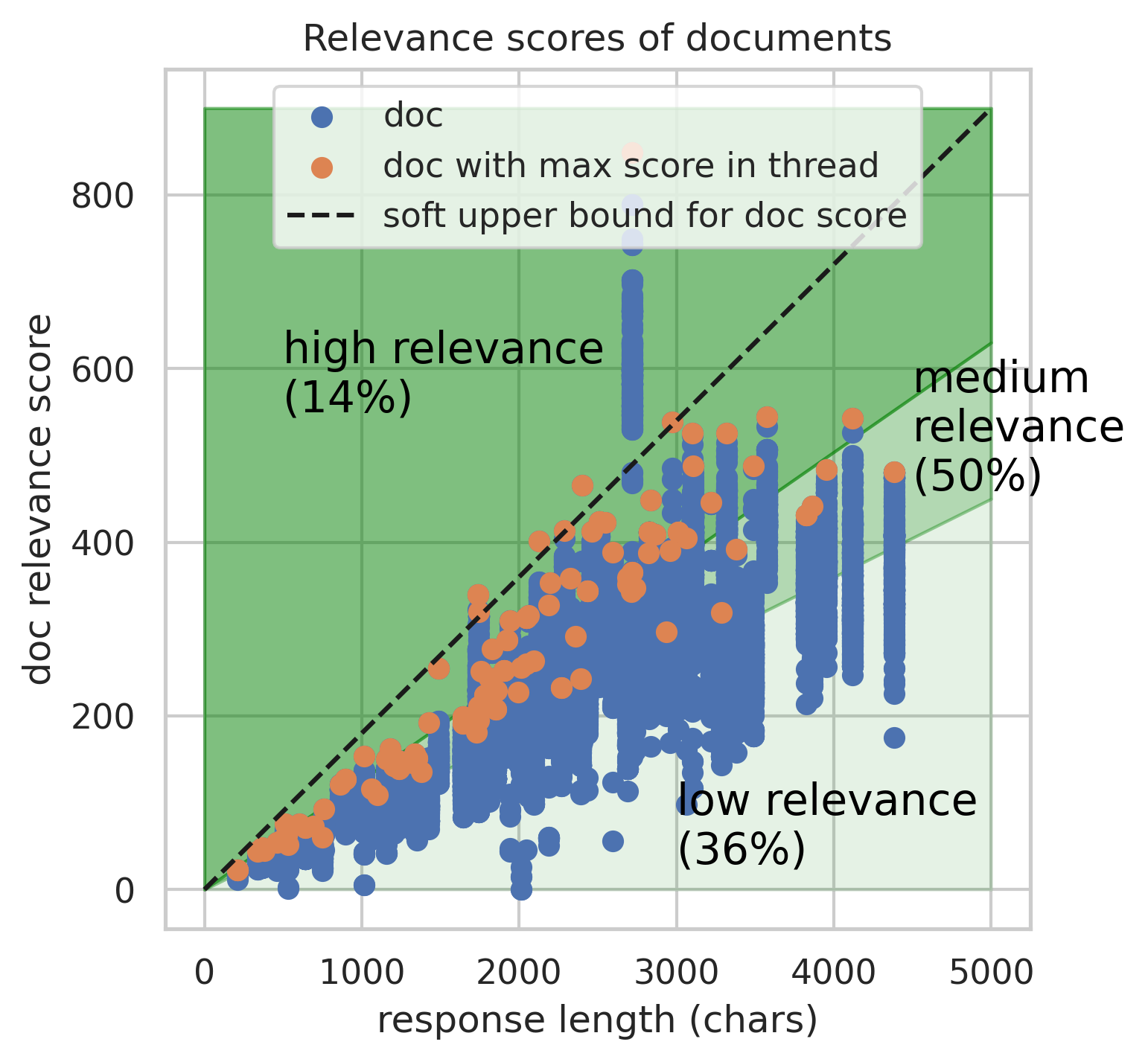}
\hfill
\includegraphics[width=0.34 \textwidth]{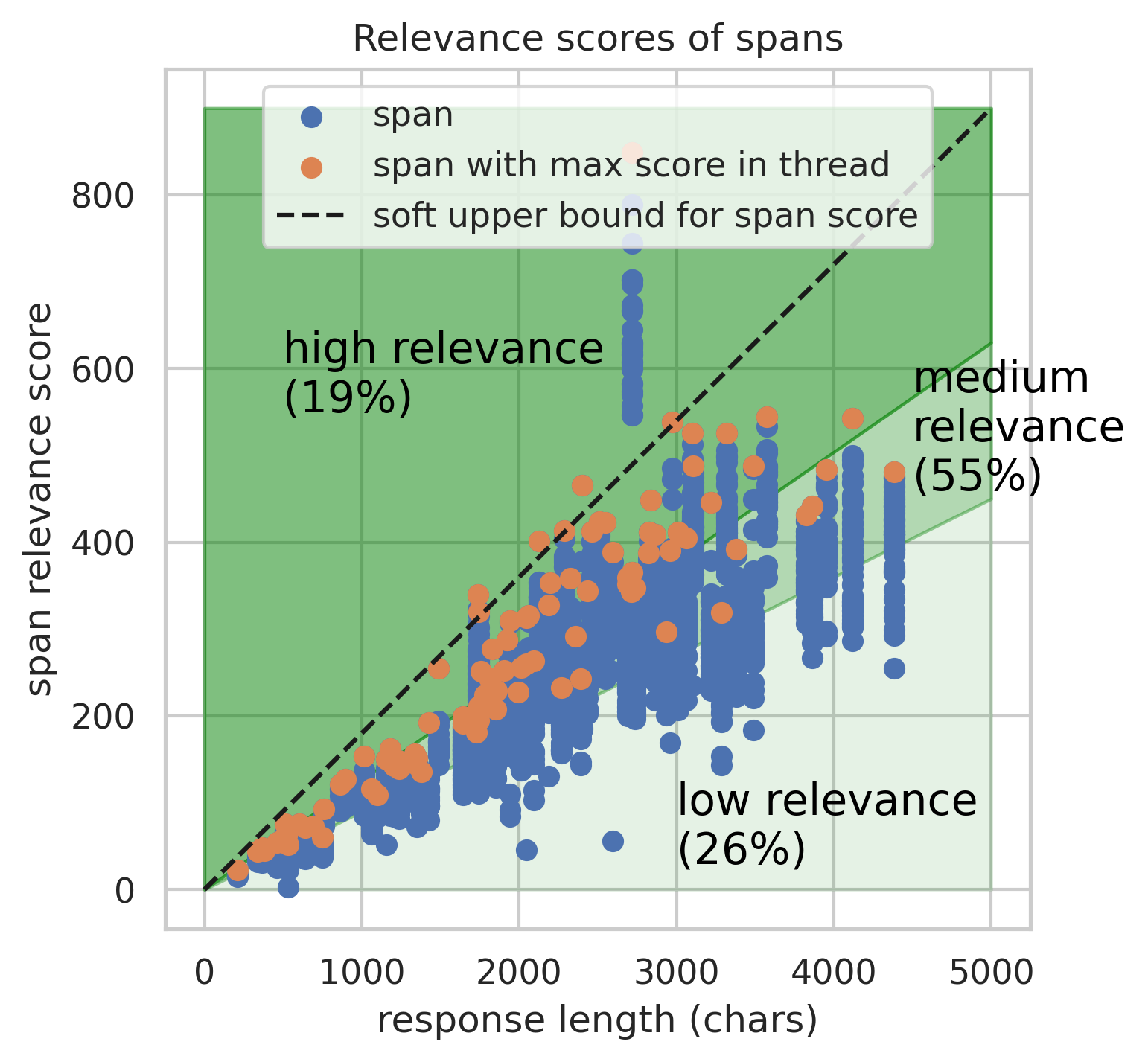}
\vspace{-8pt}
\caption{
    Statistics of spans and documents outputted by \methodname{}.
    \textbf{Left:} The spans selected are relatively long, with a mean length of 10.4 tokens.
    \textbf{Middle:} The max attainable BM25 relevance score of a document is roughly proportional to the response length, so we make the score thresholds proportional to the response length.
    \textbf{Right:} The relevance score of a span is the max score among its corresponding documents.
}
\vspace{-12pt}
\label{fig:stat}
\end{figure*}

\subsection{Benchmarking Inference Latency}
\label{sec:benchmarking}

We host the inference pipeline on a CPU-only node in the Google Cloud Platform.
The node has 64 vCPUs and 256GB RAM, and we store the index files on 40TB of SSD disks.
See App.~\S\ref{sec:production_more} for a detailed description of our production system.

We empirically benchmark the latency of the most compute-intensive part of our inference pipeline: steps 1--3, which include computing maximal matching spans and retrieving document snippets.
We collect 98 conversations from internal usage of OLMo models in the Ai2 Playground, and send them to \methodname{}.
On average, each LM response has 458 tokens, and the \methodname{} inference latency per query is \textbf{4.46 seconds}. 
This is in line with our disk I/O analysis in App.~\S\ref{sec:production_more}.
The low inference latency allows us to present \methodname{} results to users in real time and offer a smooth user experience.

\section{Analyses}
\label{sec:relevance}

We analyze the some properties of the spans and documents outputted by \methodname{}, using the same 98 conversations as in \S\ref{sec:benchmarking}.

\tightparagraph{Length of spans.}
We report the length of spans given after step 2 (filtering for long and unique spans, before merging).
The spans have a mean length of 10.4 tokens and a median of 10 tokens.
\autoref{fig:stat} (left) shows the distribution of span lengths.
This tells us that there are many long pieces of text shared between the LM output and its training data, which are revealed by \methodname{}.

\tightparagraph{Relevance score of documents.}
To improve user experience, \methodname{} reranks the retrieved documents by relevance to the LM output using BM25. 
We found that the maximum attainable BM25 score is roughly capped by 0.18 times the number of characters in the LM output (\autoref{fig:stat}, middle), so we normalize the BM25 scores by this coefficient.
After normalization, we bucket the scores as follows: $\ge 0.7$ is high relevance, between 0.5 and 0.7 is medium relevance, and $< 0.5$ is low relevance.
We empirically found these thresholds to be aligned with human expectations, and this puts 14\% of documents as high relevance.
We use the same normalization and thresholds for span scores (\autoref{fig:stat}, right), rendering 19\% of spans as high relevance.

\tightparagraph{Validating the relevance rankings.}
We conducted a study to evaluate the relevance level of the top displayed documents to the LM output.
We first composed a rubric for scoring document relevance on a 0--3 scale (App.~\autoref{tab:relevance_rubrics}, left), and asked a human expert to annotate the top-5 displayed documents for each conversation according to this rubric. 
This human evaluation round was done with \methodname{} results under a different hyperparameter setting than our final setting, and we later improved the setting under the guidance of LLM-as-a-Judge evaluation. 
The first document displayed in each conversation received an average relevance score of 1.90 (roughly meaning ``being on the same topic as the LM output), 
and the top-5 documents scored an average of 1.43 (App.~\autoref{tab:relevance_eval}). 
We then switched to LLM-as-a-Judge evaluation \citep{Zheng2023JudgingLW} with gpt-4o, 
and found that it mostly agrees with human evaluation (with a Spearman correlation coefficient of 0.73).
LLM-as-a-Judge assigned slightly lower scores overall, with average scores of 1.73 and 1.28 on first and top-5 documents, respectively.
We then used LLM-as-a-Judge to guide the tuning of several hyperparameters in \methodname{}, 
and our final setting achieved average LLM-as-a-Judge scores of 1.82 on first documents and 1.50 on top-5 documents.
See App.~\S\ref{sec:relevance_more} for additional details on relevance evaluation and hyperparameters tuning.

\tightparagraph{Training stage of retrieved documents.}
Among the retrieved documents, we found the vast majority (96.7\%) belong to the pre-training data, 0.9\% belong to the mid-training data, and 2.4\% to the post-training data.
Among post-training, 0.9\% are from the SFT data, 1.5\% are from the DPO data, and none are from the RLVR data.
We note that this distribution heavily depends on the topic of the conversation: for example, a math-heavy LM output may result in more documents retrieved from SFT and RLVR datasets.

\section{Case Studies}
\label{sec:usage}

\begin{figure}[!t]
\centering
\begin{subfigure}[t]{\linewidth}
\includegraphics[width=\linewidth]{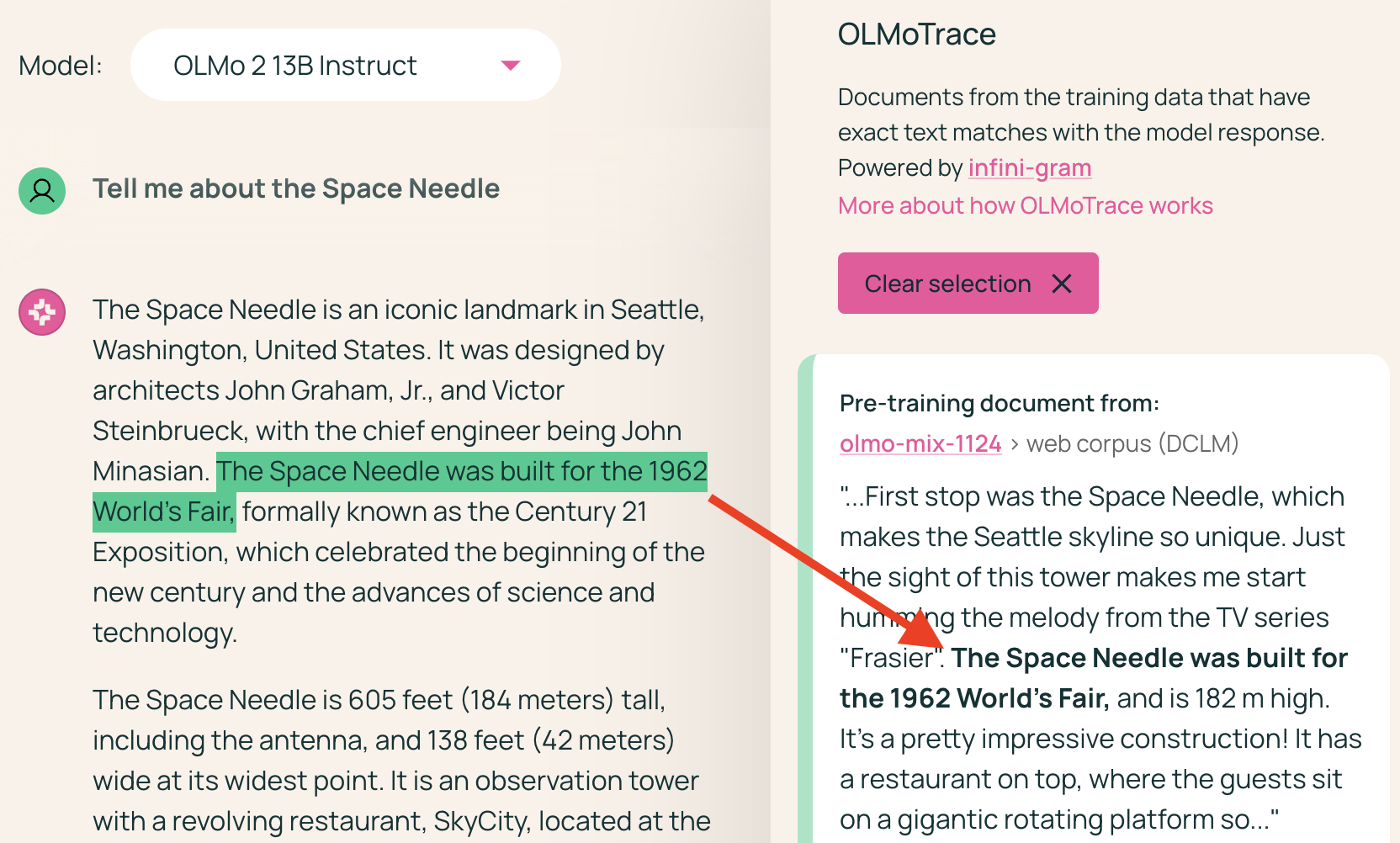}
\caption{
    \textbf{Fact checking:} Inspecting the document (and its source URL) helps verify the factual claim made in the span.
}
\end{subfigure}
\\
\vspace{4pt}
\begin{subfigure}[t]{\linewidth}
\includegraphics[width=\linewidth]{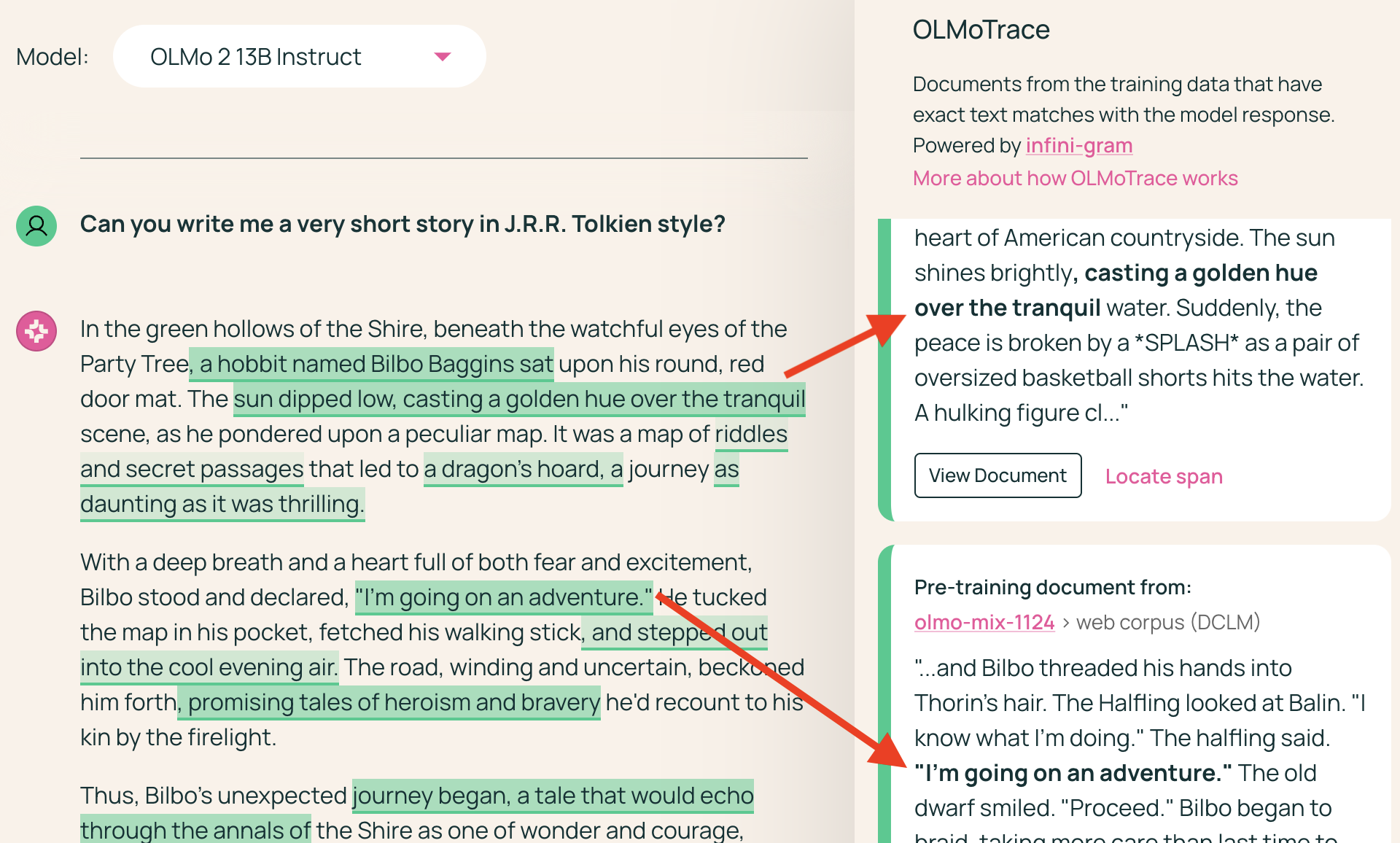}
\caption{
    \textbf{Tracing ``creative'' expressions:} Matching spans reveal potential source of LM-generated ``creative'' expressions. 
}
\end{subfigure}
\\
\vspace{4pt}
\begin{subfigure}[t]{\linewidth}
\includegraphics[width=\linewidth]{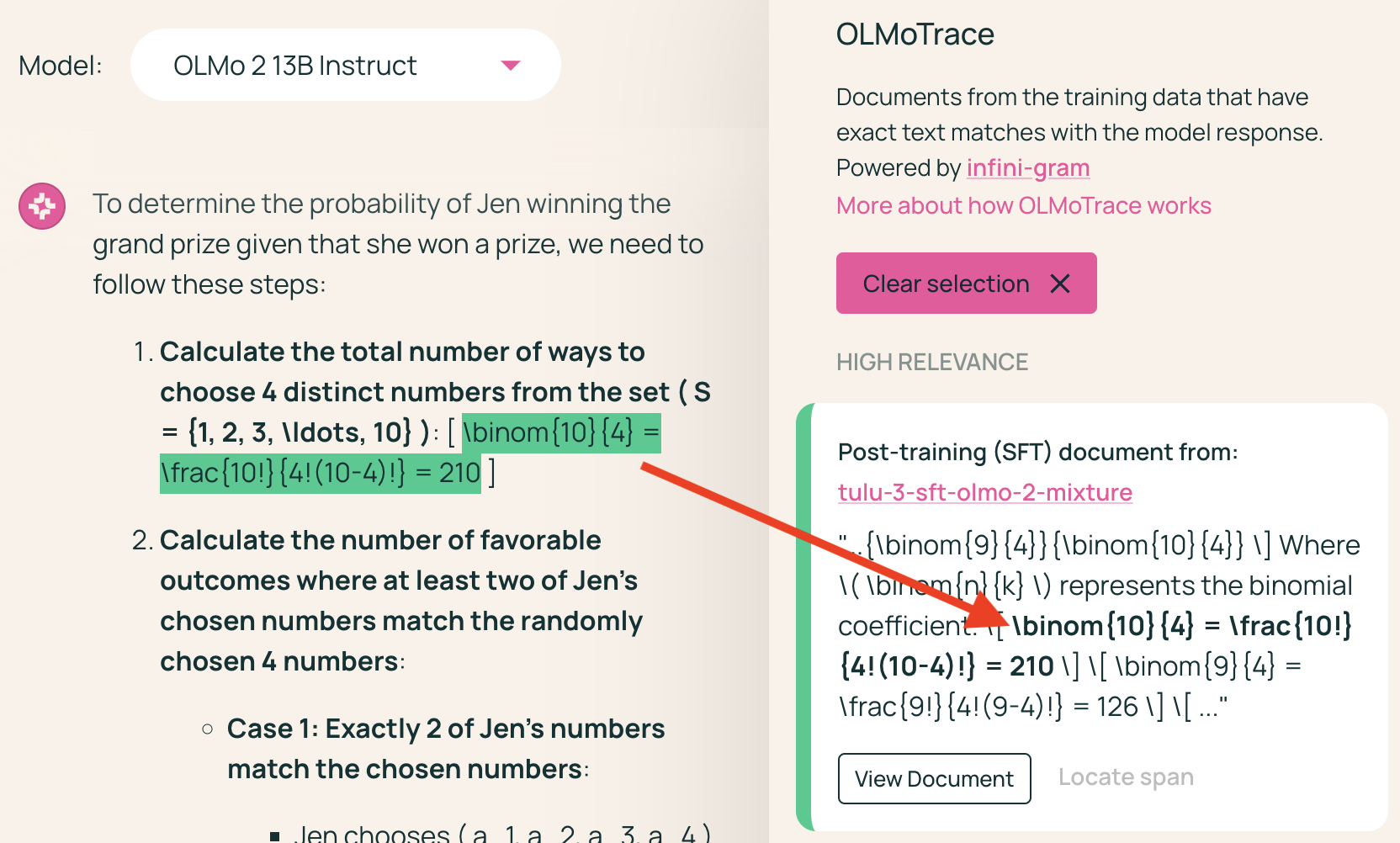}
\caption{
    \textbf{Tracing math capabilities:} Arithmetics carried out by LMs can be traced verbatim to their training data.
}
\end{subfigure}
\vspace{-7pt}
\caption{
    Example use cases of \methodname{}.
    In (a) and (c), one span has been selected to inspect its enclosing documents; the selected span is colored with solid green while other span highlights are hidden.
}
\vspace{-12pt}
\label{fig:usage}
\end{figure}

We envision that researchers and the general public can use \methodname{} in many ways to understand the behavior of LMs.
Below we discuss three example use cases, and we invite the community to explore additional ones.

\tightparagraph{Fact checking.}
If the LM states a fact, users may be able to fact-check the statement against its training data. 
In \autoref{fig:usage}(a), OLMo outputs ``\textit{\ul{The space needle was built for the 1962 World Fair,}}''.
\methodname{} highlights this span of tokens as it appears verbatim in the training data and shows the corresponding documents (the screenshot captured one of the ten documents). 
For most documents from the pretraining data (like this one), users can click on the ``View Document'' button and find the URL to the original webpage where this document was crawled.

We note that inspecting the document context and source can help users make a more informed judgment on the factuality of the statement, as words can be misleading out of context, and some web sources may be unreliable.

\tightparagraph{Tracing ``creative'' expressions.}
While LMs can be creative in piecing expressions together in new ways, seemingly novel expressions 
may not be truly new, as LMs may have learned them during training.
In such cases, \methodname{} reveals the potential source of LM-generated expressions.
In \autoref{fig:usage}(c), OLMo outputs a story in the Tolkien style, and \methodname{} highlights verbatim matches with the training data, e.g., ``\textit{\ul{I'm going on an adventure}}'' matches with the shown document, which is a fan fiction about the Hobbits.

\tightparagraph{Tracing math capabilities.}
\methodname{} helps understanding how LMs learned to carry out arithmetic operations and solve math problems.
In \autoref{fig:usage}(d), OLMo correctly answers Problem 4 from the AIME 2024 I exam (a combinatorics problem).
\methodname{} shows that the calculation step, ``\textit{\ul{\textbackslash{}binom\{10\}\{4\} = \textbackslash{}frac\{10!\}\{4!(10-4)!\} = 210}}'' appears verbatim in the post-training dataset.
\section{Related Work}
\label{sec:related}

\tightparagraph{Comparison with RAG.}
Retrieval-augmented generation (RAG) systems retrieve relevant documents from a database and condition the LM generation on the retrieved documents.
Examples of them include Bing Chat, Google AI Overview, and Perplexity AI.
Despite looking similar, \methodname{} is fundamentally different from RAG: \methodname{} retrieves documents \textit{post-hoc} and does not intervene with the LM generation. 
The purpose of retrieval in \methodname{} is to show the connection between an LM's output and its training data, not to improve the generation itself. 

\tightparagraph{Comparison with search engines.}
Traditional search engines (e.g., Google) retrieve documents from their web index. 
\methodname{} retrieves matches in an LM's training data, which is more suitable to use for understanding the data origin of LM behaviors.

\tightparagraph{Tracing LM generation into training data.}
One classical approach to trace LM generation is using influence functions \citep{Koh2017UnderstandingBP,Han2020ExplainingBB,Han2022ORCAIP}, which leverage gradient information to find influential training examples for a given test example. 
While effective on a small scale, influence functions are intractable for trillion-token training data due to their high computational cost.
Our work takes a different approach: we directly retrieve similar training examples by lexical overlap, with the heuristic that such training examples are likely to be influential for the given output.

\tightparagraph{Other types of tracing.} 
\citet{Khalifa2024SourceAwareTE} train LMs to cite documents from the pretraining data, which is an intervention on the training process of LMs.
Some work traces LM behavior into sources other than the training data.
\citet{Huang2024TrainingLM} extend RAG to have LMs cite retrieved documents provided in-context, whereas \citet{Chuang2025SelfCiteSA} train LMs to cite content from the long context provided to the LM at inference time.
\citet{Gao2022RARRRA} retrieve supporting evidence for LM generations from Google Search; the Gemini App has a ``double-check response'' feature that highlights parts of the LM response and shows similar results from Google Search, which is updated in real time and thus not identical to Gemini's training data, making it less useful for scientific exploration. 

\clearpage
\section*{Limitations}
\label{sec:limitations}

\methodname{} finds lexical, verbatim matches between an LM's output and its training data.
The retrieved documents should not be interpreted as having a causal effect on the LM output, or as supporting evidence or citations for the LM output.

\tightparagraph{Mitigating social and legal risks.}
\methodname{} can make potentially problematic contents in the LM training data more easily exposed.
We conducted an internal red-teaming effort and implemented mitigation measures based on the findings.
We focused on three aspects: copyright, PII (personal identifiable information), and toxicity.
For copyright, we were able to make \methodname{} show documents with news articles or song lyrics, while we did not see any copyrighted book; we offer a takedown request form for copyright holders to fill out in case they identify documents infringing their copyright, and we implemented an efficient way to take down documents in the \infinigram{} engine so that we don't need to re-index the full training data.
For PII, we were unable to find any PII data in \methodname{} results, and we implemented a regex-based filter to block documents with PII.
For toxicity, text moderation is already implemented in Ai2 Playground to filter user prompts, and we do not add further filtering.

\section*{Acknowledgements}
\label{sec:ack}

We would like to thank Alisa Liu, Valentina Pyatkin, Dany Haddad, Shannon Zejiang Shen, Ziqi Ma, and other members of Ai2, H2lab and Xlab for sharing their valuable feedback.
This work was funded in part by NSF IIS-2044660.

\section*{Author Contributions}

\tightparagraph{Core contributors:} Jiacheng Liu, Taylor Blanton

\tightparagraph{Research:} Yanai Elazar, Sewon Min, Luca Soldaini, Dirk Groeneveld, Rock Yuren Pang

\tightparagraph{Engineering:} YenSung Chen, Arnavi Chheda-Kothary, Huy Tran, Byron Bischoff, Eric Marsh

\tightparagraph{Design:} Cassidy Trier, Aaron Sarnat, Jenna James, Jon Borchardt

\tightparagraph{UX:} Bailey Kuehl, Evie Cheng

\tightparagraph{PM:} Karen Farley, Sruthi Sreeram, Taira Anderson

\tightparagraph{Legal:} Will Smith, Crystal Nam

\tightparagraph{Comms:} David Albright, Carissa Schoenick

\tightparagraph{Advising:} Jesse Dodge, Ali Farhadi, Hannaneh Hajishirzi, Yejin Choi, Sophie Lebrecht, Noah A. Smith, Pang Wei Koh

\bibliography{custom}

\clearpage
\appendix

\section{More Screenshots of \methodname{}}

\begin{figure*}[!b]
\centering
\includegraphics[width=0.49 \textwidth]{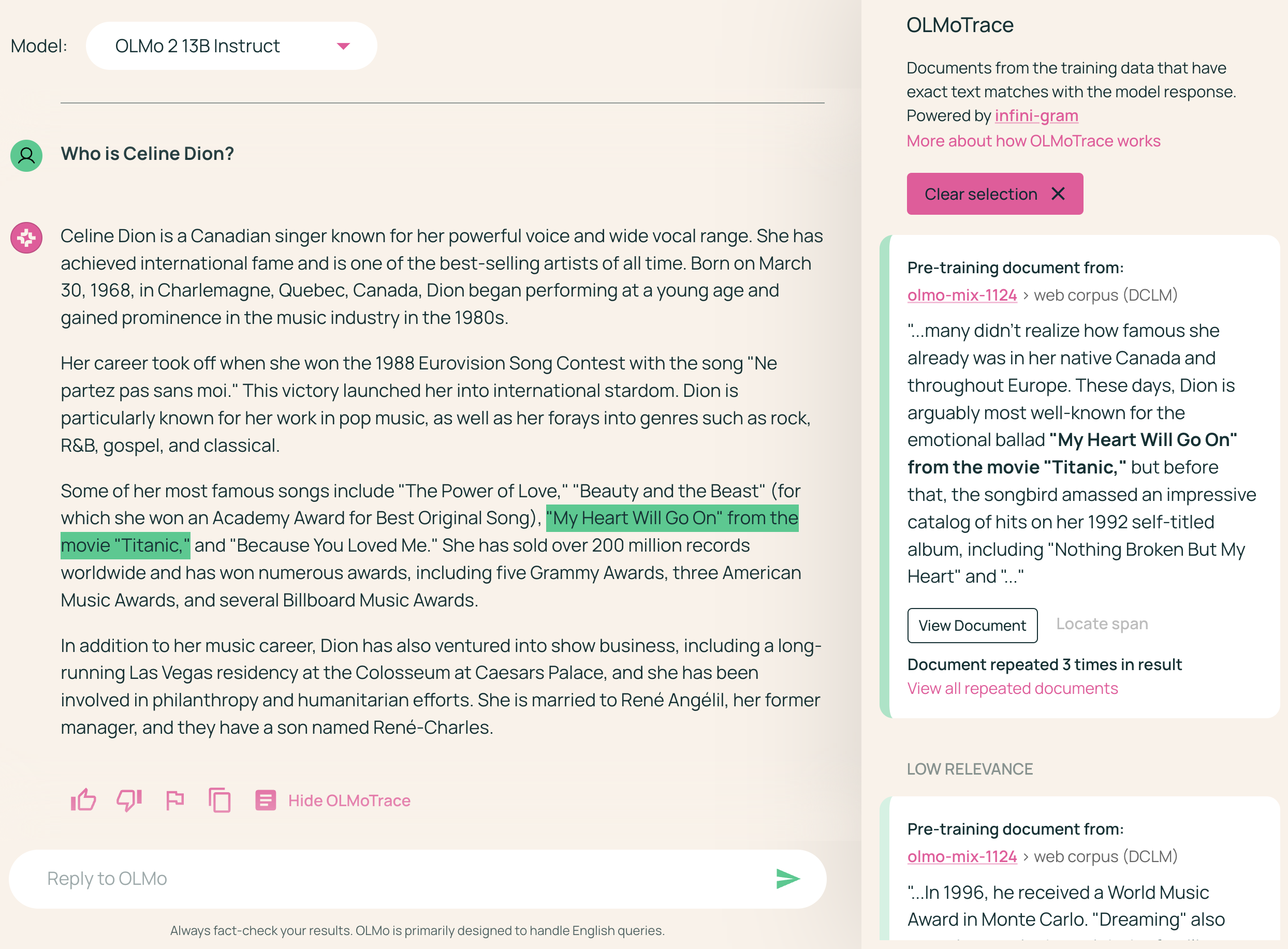}
\hfill
\includegraphics[width=0.49 \textwidth]{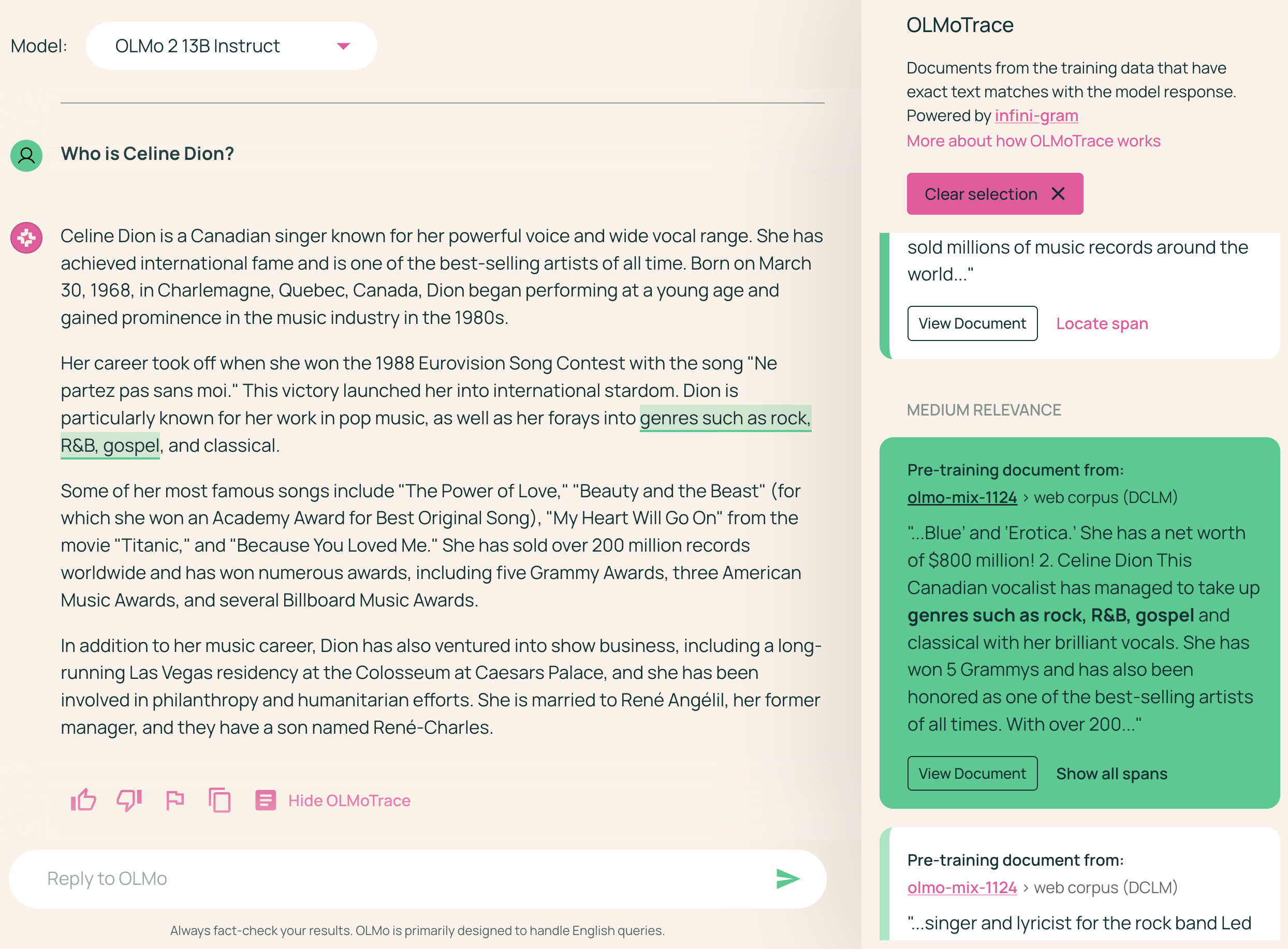}
\vspace{-8pt}
\caption{
    Screenshots on interacting with \methodname{} on Ai2 Playground.
    \textbf{Left:} When user clicks on a highlighted span, the document panel is filtered to only present documents enclosing the selected span.
    \textbf{Right:} When user clicks the ``Locate Span'' button on a document, the span highlights will narrow down to those enclosed in the selected document.
    Clicking on the same place again or the ``Clear Selection'' button will lead back to showing all spans and documents (\autoref{fig:playground}, left).
}
\label{fig:playground_more}
\end{figure*}

\autoref{fig:playground_more} is an extension of \autoref{fig:playground} and shows screenshots of \methodname{} when user interacts with the UI.




\section{Production System Setup}
\label{sec:production_more}

We host the production system of \methodname{} on Google Cloud Platform.
We store the \infinigram{} index files on \texttt{pd-balanced} SSD disks with up to 80,000 read IOPS per VM.
To achieve the maximum IOPS, we mount the disks to an N2 VM with 64 vCPUs.
We keep the index files on disk for inference to avoid needing an unrealistic amount of RAM, and allocate 256GB RAM for the VM to fit the fully-materialized page tables of the mmap'ed index files (0.2\% the full file size).
To enhance system availability and throughput, we keep 2 VM replicas and multi-mount the same disks to both VMs, and we keep \methodname{} processing in separate workers.

In the \infinigram{} engine, we turn off prefetching (setting all prefetch depth to 0) because it would slow down the overall inference.
(Prefetching reduces the latency of single query at the cost of performing more disk read ops speculatively, which is not beneficial when disk I/O throughput is the bottleneck.)
We also implemented a batched version of \textsc{GetDocByPtr} query to retrieve multiple training documents in parallel and reduce latency.

\tightparagraph{Disk I/O analysis.}
Here we compute the number of random disk reads needed in the span computation step.
For each beginning token position in the LM output, we need to find its longest matching prefix which means 2 \textsc{Find} queries; effectively this only counts as 1 \textsc{Find} query because most disk reads are shared and cached.
Each \text{Find} takes 2 binary searches over the SA, but our implementation combines them into 1 binary search.
Each binary search takes $\log N \approx 40$ steps, where each step takes 2 disk reads -- one on the SA and one on the text corpus.
In practice we partition the training data into 12 shards, so multiply the number of disk reads by 12.
This means for each token in the LM output, we need $40 \times 2 \times 12 = 960$ disk reads.
Given that our disks have 80,000 IOPS, \methodname{} can processes, for example, a 100-token LM output within 1.2 seconds.

\section{More Details on Document Relevance Evaluation}
\label{sec:relevance_more}

\begin{table*}[!t]
\centering
\setlength{\tabcolsep}{4pt}
\resizebox{0.61 \linewidth}{!}{%
\begin{tabular}{l p{320pt}}
\toprule
\textbf{Score} & \textbf{Description} \\
\midrule
0 & The snippet or context of the snippet is about a different topic than the query and model response (though possibly semantically similar): \newline For example, for the query breast cancer symptoms, give a 0 to: \newline \quad A snippet about heart attack symptoms  – wrong topic \newline \quad A snippet about brain cancer symptoms –  may not necessarily apply to breast cancer symptoms \\
\midrule
1 & The snippet or context of the snippet is about a broader topic than the query and model response, or is potentially relevant but there’s not enough information: \newline For example, for the query breast cancer symptoms, give a 1 to: \newline \quad A snippet about cancer in general – missing key specifics of symptoms \\
\midrule
2 & The snippet or context of the snippet is on the right topic of the query and model response, but is in a slightly different context or is too specific to fit the exact query: \newline For example, for the query breast cancer symptoms, give a 2 to: \newline \quad A snippet referring a breast cancer treatment side effect \\
\midrule
3 & The snippet or context of the snippet is about a subject that is a direct match, in topic and scope, of the most likely user intent for the query and model response: \newline For example, for the query breast cancer symptoms, give a 3 to: \newline \quad A snippet discussing a symptom specific to breast cancer \\
\bottomrule
\end{tabular}
}%
\hfill
\resizebox{0.38 \linewidth}{!}{%
\begin{tabular}{p{230pt}}
\toprule
\textbf{LLM-as-a-Judge Prompt} \\
\midrule
You will be given a user prompt, a model's response to the prompt, and a retrieved document. Please rate how relevant the document is to the prompt and model response. Rate on a scale of 0 (not relevant) to 3 (very relevant). Respond with a single number, and do not include any other text in your response.
\newline \newline
Rubric for rating:
\newline
0: The document is about a different topic than the prompt and model response.
\newline
1. The document is about a broader topic than the prompt and model response, or is potentially relevant but there's not enough information.
\newline
2. The document is on the right topic of the prompt and model response, but is in a slightly different context or is too specific.
\newline
3. The document is about a subject that is a direct match, in topic and scope, of the most likely user intent for the prompt and model response.
\newline \newline
Prompt: \{prompt\}
\newline
Model response: \{response\}
\newline
Retrieved document: \{document\}
\end{tabular}
}%
\vspace{-8pt}
\caption{
    \textbf{Left:} Rubrics for document relevance evaluation.
    \textbf{Right:} Prompt for automatically evaluating document relevance with LLM-as-a-Judge.
}
\vspace{-8pt}
\label{tab:relevance_rubrics}
\end{table*}
\begin{table*}[!t]
\centering
\setlength{\tabcolsep}{4pt}
\resizebox{\linewidth}{!}{%
\begin{tabular}{l r r r r}
\toprule
& \textbf{Avg score} & \textbf{Avg score} & \textbf{\% relevant} & \textbf{\% relevant} \\
\textbf{Setting} & \textbf{(1st doc)} & \textbf{(top-5 docs)} & \textbf{(1st doc)} & \textbf{(top-5 docs)} \\
\midrule
our final setting & \textbf{1.82} & \textbf{1.50} & 63.3\% & \textbf{55.1\%} \\
+ BM25 doc reranking only considers LM response (no user prompt) & 1.78 & 1.49 & 62.2\% & 54.5\% \\
\hspace{4pt} + shorten doc context length to 100 tokens & 1.74 & 1.44 & \textbf{64.3\%} & 52.9\% \\
\hspace{8pt} + span ranking w/ length & 1.56 & 1.37 & 57.1\% & 49.4\% \\
\hspace{12pt} + drop spans w/ frequency >10 & 1.73 & 1.28 & 62.2\% & 47.0\% \\
\midrule
\hspace{16pt} + switch to human annotator & 1.90 & 1.43 & 63.0\% & 46.2\% \\
\bottomrule
\end{tabular}
}%
\vspace{-8pt}
\caption{
    Evaluating the relevance level of top documents displayed by \methodname{}.
    Avg score is on a likert scale of 0-3, where 0 is ``unrelated'' and 3 is ``highly relevant''.
    For \% relevant, we consider a document as relevant if it gets a score of 2 or 3.
    We use LLM-as-a-Judge with gpt-4o-2024-08-06, except in the last row where we collect annotation from a human expert.
}
\vspace{-12pt}
\label{tab:relevance_eval}
\end{table*}

For human evaluation, we used the rubric in \autoref{tab:relevance_rubrics} (left).
For LLM-as-a-Judge evaluation, we used the prompt in \autoref{tab:relevance_rubrics} (right), which closely follows the rubric, and gpt-4o-2024-08-06 as the judge model.

\autoref{tab:relevance_eval} shows the evaluation results.
We report 4 metrics: average score among the first and top-5 displayed documents, and the percentage of relevant documents among the first and top-5 displayed documents.
We report different settings in reversed chronological order.
For the last row, we used an early hyperparameter setting of \methodname{} with human evaluation, and for the second-last row we used the same hyperparameter setting but switched to LLM-as-a-Judge.
The early hyperparameter setting differs from our final setting in that:
\begin{enumerate}[leftmargin=16pt, itemsep=-4pt, topsep=4pt]
\item Before step 2, it dropped maximal matching spans that appear more than 10 times in the training data (i.e., frequency >10);
\item In step 2, it ranked the spans by descending length instead of ascending span unigram probability;
\item When reranking documents in step 5, the BM25 scorer only considered a context length of 100 tokens around the span instead of 500;
\item The BM25 scorer only considered the LM response and did not consider the user prompt.
\end{enumerate}
We tuned LLM-as-a-Judge so that it has high agreement and roughly matched statistics with the human evaluation, and our selection of model (gpt-4o-2024-08-06) and prompt (\autoref{tab:relevance_rubrics}, right) was the best combination we reached.

We then incrementally adjusted the hyperparameter settings in \methodname{} and measured the document relevance with LLM-as-a-Judge.
The first change we made is to no longer drop maximal matching spans that appear more than 10 times in the training data.
The dropping was due to a limitation in the early version of our system, and we thought this would lead to incomplete results (many documents are duplicated more than 10 times in the pre-training data) and decided to not drop any maximal matching spans according to frequency.
Not dropping spans decreased the metrics on the first displayed documents, but increased the metrics on the top-5.
Subsequently, we incrementally flipped item 2, 3, and 4 in the above change list, and with every change applied, the overall document relevance metrics improved (with a small exception on \% relevant among first displayed documents).
Our final setting achieved an average relevance score of 1.82 among the first displayed documents, and 1.50 among the top-5 documents, according to LLM-as-a-Judge.

\end{document}